\documentclass[letterpaper, 10 pt, conference]{ieeeconf}  

\IEEEoverridecommandlockouts                              

\usepackage{multirow, makecell}
\usepackage{hhline}
\usepackage{array, booktabs}
\usepackage{xcolor}

\usepackage{graphicx}
\usepackage{enumerate}
\usepackage{epsfig} 
\usepackage{subcaption}
\usepackage{amsmath} 
\usepackage{amssymb}  

\usepackage{bm}
\usepackage{multirow, makecell}
\usepackage[linesnumbered,ruled,vlined]{algorithm2e}
\usepackage{cite}
\usepackage{hyperref}
\usepackage{cleveref}
\usepackage{algpseudocode}
\usepackage[font={small}]{caption}
\usepackage{color}

\usepackage{floatrow}
\newfloatcommand{capbtabbox}{table}[][\FBwidth]

\newcommand{\comment}[1]{}

\newcommand{\ie}{{\em i.e. }}

\newcommand{\bmm}{\mathbf{m}}
\newcommand{\bhm}{\hat{\mathbf{m}}}
\newcommand{\btm}{\tilde{\mathbf{m}}}

\newcommand{\bM}{\mathbf{M}}

\newcommand{\bX}{\mathbf{X}}

\newcommand{\bhM}{\hat{\mathbf{M}}}

\newcommand{\mM}{\mathcal{M}}

\newcommand{\mL}{\mathcal{L}}

\newcommand{\mS}{\mathcal{S}}

\usepackage{blindtext}

\title{\LARGE \bf
Grasp-Oriented Fine-grained Cloth Segmentation \\ without Real Supervision
}

 \author{Ruijie Ren$^{1,*}$, Mohit Gurnani Rajesh$^{2,*}$, Jordi Sanchez-Riera$^{1}$, Fan Zhang$^{2}$, Yurun Tian$^{2}$, Antonio Agudo$^{1}$\\
 Yiannis Demiris$^{2}$, Krystian Mikolajczyk$^{2}$ and Francesc Moreno-Noguer$^{1}$  
 \thanks{$^{1}$Authors are with the Institut de Robòtica i Informàtica Industrial, CSIC-UPC.  C/ Llorens i Artigas 4-6, 08028, Barcelona, Spain.
 {\tt\footnotesize [rren, jsanchez, aagudo, fmoreno]@iri.upc.edu}}%
 \thanks{$^{2}$Authors are with the Imperial College London, UK.
 {\tt\footnotesize [mohit.rajesh-gurnani18, f.zhang, y.tian, y.demiris, k.mikolajczyk]@imperial.ac.uk}}
  \thanks{$^*$ First two authors contributed equally}}

\begin{document}

\maketitle

\thispagestyle{empty}
\pagestyle{empty}

\begin{abstract}
Automatically detecting graspable regions from a single depth image is a key ingredient in cloth manipulation. The large variability of cloth deformations has motivated most of the current approaches to focus on  identifying specific grasping points rather than semantic parts, as the appearance and depth variations of local regions are smaller and easier to model than the larger ones. However, tasks like  cloth folding or assisted dressing require recognising larger segments,  such as semantic edges  that carry more information  than points. The first goal of this paper is therefore to tackle the problem of fine-grained region detection in deformed clothes using only a depth image. As a proof of concept, we implement an approach for T-shirts, and define up to 6 semantic regions of varying extent, including edges on the neckline, sleeve cuffs, and hem, plus top and bottom grasping points. We introduce a U-net based network to segment and label these parts. The second contribution of our work is concerned with the level of supervision that we require to train the proposed network. While most approaches learn to detect grasping points by combining real and synthetic annotations, in this work we defy the limitations of the synthetic data, and propose a multilayered domain adaptation (DA) strategy that does not use real annotations at all. We thoroughly evaluate our approach on real depth images of a T-shirt annotated with fine-grained labels. We show that training our network solely with synthetic data and the proposed DA yields results competitive with models trained on real data.


\end{abstract}

\section{INTRODUCTION}
\begin{figure}[t!]
    \centering
    \includegraphics[width=\linewidth]{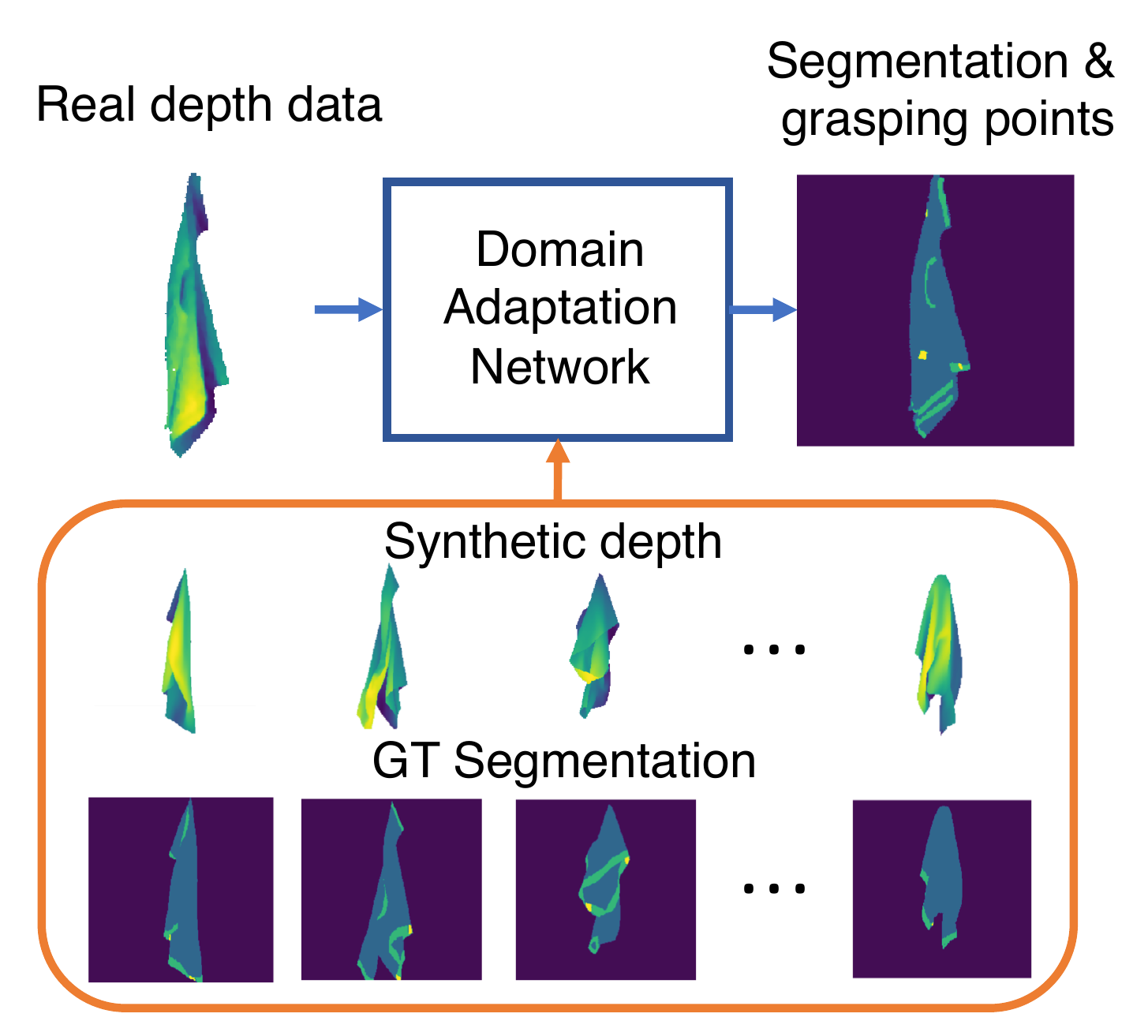}
    \caption{Overview of our approach. We introduce a pipeline for fine-grained semantic segmentation of depth maps of clothes. The segments we provide are needed for performing subsequent manipulation tasks, include grasping points and different types of semantic edges. Additionally, we propose a multi-layered domain adaptation strategy to train the proposed network with only synthetic GT labels, which can then be applied to real depth maps.}
    \label{fig:intro}
\end{figure}

Identifying specific regions for grasping  is one of the main open challenges in robotic cloth manipulation, due to the large variability of the possible geometric configurations and textures that garments exhibit. Texture variability can be surpassed by the use of depth data, although the problem is still severely ill-posed. See  the depth map of a T-shirt in Fig.~\ref{fig:intro}. Identifying semantically meaningful regions in that range image is extremely difficult even for a human. Most existing methods have therefore focused on identifying key grasping points, as small local regions tend to have more invariant geometric features than large patches~\cite{Saxena-SII-2019,Fan2020,classical,CORONA2018629,Mariolis2015,Qian2020ClothRS,Maitin-shepard10clothgrasp,Ramisa2011DeterminingWT}.  Nevertheless, many tasks in cloth manipulation would benefit from detecting larger regions that carry more information and are more suitable for the task to be performed. In particular, the semantic edges (e.g. neckline, sleeve cuffs and hem in a T-shirt), are very useful for tasks like folding or dressing people.

The first contribution of this work is therefore a method to perform fine-grained edge detection on highly crumpled clothes. We focus on the specific case of a T-shirt, although, the approach is generalizable to other garments. We define up to 6 semantic labels of different types and extents, including the full body, 3 edge types, and 4  grasping points.  We formulate this task as semantic segmentation, and adapt a U-Net architecture with reversed gradient layers, that given input depth images, can provide the semantic labels of each category (even grasping points are treated as regions).

The second contribution of our work is addressing the amount of supervision required to train our network. Existing approaches use either a training set of real depth maps annotated with ground truth grasping points 
or a combination of those with synthetic annotations~\cite{Saxena-SII-2019,Fan2020,CORONA2018629,Mariolis2015,Maitin-shepard10clothgrasp}. In this work, motivated by the difficulty of collecting and annotating real ground truth depth maps with fine-grained edge labels for training, we explore the limits of relying exclusively on supervision from  synthetically generated data. For this purpose, we synthesize and annotate several thousands of samples of a T-shirt hanging under gravity from random points. Additionally, we also create a dataset of real depth maps pseudo-annotated with the fine-grained labels for testing our methods. We then investigate several training alternatives. More specifically, we propose a Multi-layer Domain Adaptation approach in which the transformations over feature maps extracted from  synthetic input depth maps are ruled by the reverse gradient of an adversarial loss computed from non-annotated synthetic/real samples. A thorough evaluation shows that this scheme achieves results competitive with architectures trained on the pseudo-annotated real samples.

In summary, the main contributions of our work are the following:
\begin{itemize}
\item We explore, for the first time, the problem of fine-grained edge segmentation in depth maps of highly deformed clothes.
\item We explore the limits of domain adaptation strategies that leverage uniquely on supervision from synthetic annotations.
\item We generate large and realistic synthetic data and collect a mid-size real dataset of deformed T-shirts which we annotated with  edge labels and grasping points. This dataset can be used either for finetuning synthetically trained networks or for evaluation, and will be made publicly available together with the proposed model.

\end{itemize}



\section{RELATED WORK}

There have been multiple works that focus on manipulating highly deformable objects such as clothes. Most of these works concentrate on finding suitable grasping points either for towels \cite{Ramisa2011DeterminingWT, Gabas2017} or for more structured clothes like t-shirts, pants or sweaters \cite{Ramisa-AAI-2014, Li-icra-2015}. Typically, after capturing data with a depth sensor device,  early methods concentrate on finding geometric cues \cite{Sun-icra-2015, Maitin-shepard10clothgrasp, classical} (i.e. cloth folds and wrinkles, cloth corners, etc.) or classify cloth deformation to indirectly infer the grasping points \cite{Stria2018ClassificationOH, Kita-icra-2009}. However, these kind of approaches are difficult to use for complex clothes, as  the detected edges or other geometric cues lack semantic meaning, which needs to be compensated by using fiducial markers on the cloth \cite{Bersch2011BimanualRC} for a more reliable detection of grasping points.

 Recent methods exploit the potential of neural networks. In order to train these networks, it is necessary to use a large amount of data, that can be achieved by means of generating  synthetic datasets \cite{Saxena-SII-2019, CORONA2018629}. Unfortunately, networks trained exclusively on synthetic data have problems to generalize when using real examples. For this reason, synthetic data  is often mixed with real data which can be acquired by painting a white cloth with the desired annotation marks \cite{Qian2020ClothRS, Gabas2017}. This procedure is tedious and makes the data generation more complex  as it involves robot manipulation to obtain images and a pre-processing operations to extract the annotations. Therefore, other methods train the networks with synthetic data and later use a small amount of real examples to fine-tune the grasping point detection network \cite{Mariolis2015, Fan2020}.
In contrast, our work makes use of Domain Adaptation (DA) methodology to narrow the gap between synthetic and real data. The main advantage of using DA is that it eliminates the need of collecting and annotating large real datasets, allows the proposed network to be trained by supervision only from synthetic examples, while achieving comparable results to the methods trained with real  or a mixture of real and synthetic data.
We believe that this characteristic makes our proposed method easily generalizable to various types of garments. Moreover, unlike  the methods discussed above that focus on grasping point detection, our proposed approach is also able to detect semantically meaningful regions (e.g. neck, sleeve cuffs, hem) that can facilitate manipulating the cloth, especially in the case of occluded grasping points.

\begin{figure*}[t!]
    \centering
    \includegraphics[width=\linewidth]{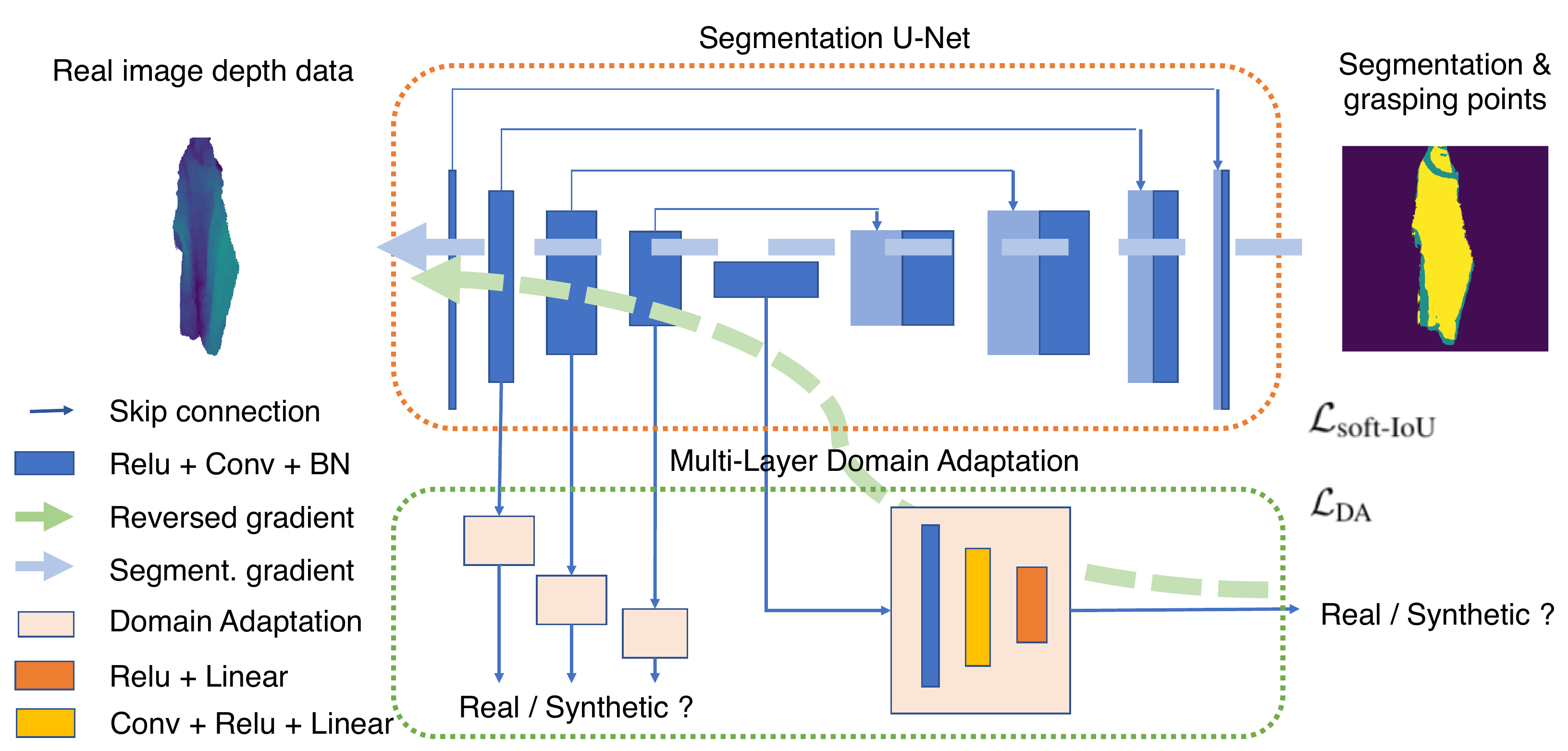}
    \caption{Our approach for fine-grained segmentation of cloth depth maps. It consists of two main branches, a U-Net (top-branch) that segments the cloth parts and a  multi-layered domain adaptation classifier (bottom) that helps to reduce the domain gap between real and synthetically generated depth maps. This strategy allows our network to generalize to real depth maps, despite being trained with synthetic data. The segmentation gradient is only computed for the segmentation loss of synthetic samples. {Unlabelled real data is only used in the Multi-Layer DA branch, in order to reduce the difference between the real/synthetic features computed by the U-Net.} }
    \label{fig:pipeline}
\end{figure*}

\section{METHOD}\label{method}
In this section we first formalise our problem, which is to design a deep learning model able to segment depth maps of clothes into semantic regions, that are tailored to perform manipulation and grasping tasks. We then describe the model we propose as well as the  training process that uses synthetically generated data.

\subsection{Problem Formulation}

Let $\bX$ be a $H\times W$ depth map of a cloth hanging under gravity from a random point. Let us also define  reference segmentation masks $\bM=\left[\bmm_1,\ldots,\bmm_C\right]$ for the $C$ cloth labels, where $\bmm_i$ is a binary mask indicating the region in the depth map that belongs to the $i$-th label. {As we shall see in the experimental section, we consider segments of very different size, from small regions defining grasping points, to elongated and large areas of semantic edges, as well as the whole body of the cloth.  We will also show that treating the grasping points as regions, and detecting them using a segmentation approach, yields improved results compared to the methods that locate them using a network regressing directly their coordinates~\cite{Fan2020}}

Furthermore, we  define $\mS^s=\left\{\bX^s_i,\bM^s_i\right\}$, $i=1,\ldots,M_s$, a set of synthetically generated pairs of depth maps and ground truth masks, and  $\mS^r=\left\{\bX^r_i,\bM^r_i\right\}$, $i=1,\ldots,M_r$, pairs of real depth maps and pseudo-ground truth masks, as described in section~\ref{sec:dataset}.

Our goal is to estimate  masks $\bhm_i$ of relevant cloth parts from a given depth map $\bX$, \ie   to learn the mapping $\mM:\bX \longrightarrow \bhM$, where $\bhM=\left[\bhm_1,\ldots,\bhm_C\right]$. In order to train $\mM$, we explore a domain adaptation learning scheme that uses the synthetically generated ground truth data  $\mS^s$, as well as the real depth maps $\bX^r_i$ without annotations. The real pseudo-ground truth masks $\bM_i^r$ will not  be used during training of our main approach.

\subsection{Model}

The architecture used in our approach is illustrated in Fig.~\ref{fig:pipeline}. It is composed of two main modules: a segmentation U-Net and a multi-layer domain adaptation classifier.

Given an input depth map $\bX$, the {\em segmentation U-Net} aims to classify every input pixel in $\bX$ into one of the $C$ cloth part categories. We implement this module following a standard convolutional U-Net architecture~\cite{ronneberger2015u}, with four encoder and four decoder blocks.

As we show in the experimental section, training the U-Net network uniquely with synthetically generated data $\mS^s$ does not generalize well to real depth maps. In order to narrow the domain gap between real and synthetic depth maps, we introduce a multi-layer domain adaptation strategy~\cite{ciga2019multi}. More specifically, the features extracted at each encoder block of the U-Net, are forwarded to multiple classifiers, which aim at distinguishing between the feature maps belonging to real ($\bX^r_i$) or synthetic ($\bX^s_i$) samples. The gradient of this domain classification loss is then reversed and backpropagated to the U-Net network that performs the segmentation~\cite{ganin2015unsupervised}. Note that by backpropagating the reverse gradient of the classification loss, we are encouraging the U-Net  encoder layers to produce features that are domain-independent, and thus, bridging the gap between the synthetic and real domains.

These domain classifiers are implemented as simple convolutional layers followed by a fully connected layer that performs a binary classification.

\subsection{Loss Functions}
We define the loss function that contains two terms, namely a semantic segmentation loss to assess the quality of the segmentation, and a  domain-adaptation loss that forces the extracted features of the real and synthetic domains to be similar. 

\vspace{1mm}
\noindent{\em Segmentation loss.} To penalize pixel-wise mask errors we use a weighted soft Intersection over Union (IoU) loss, defined as:
 \begin{equation*}
    \mathcal{L}_{\textrm{soft-IoU}} = \frac{1}{\left| C \right|}\sum_c{\frac{w_c\sum_x{\btm_{c}(x) \bmm_{c}(x)}}
    {w_c\sum_x{\btm_{c}(x)+\bmm_{c}(x)-\btm_{c}(x) \bmm_{c}(x)   }}}
\end{equation*}
where $\btm_{c}(x)$ is the prediction score at the image location $x$ for class $c$, and $\bmm_{c}(x)$ is the ground truth, which is a delta function centered at  the correct label. Note that, we added a weight factor $w_c$ for each label to deal with the label imbalance problem.

\vspace{1mm}
\noindent{\em Domain Adaptation loss.} DA loss, which we denote as $\mL_{\textrm{DA}}$,  corresponds to the error produced in a binary classification problem, with classes ``real" or ``synthetic". We compute it as a standard cross-entropy loss.

\vspace{1mm}
\noindent{\em Total loss.} We define the total loss as a linear combination of the two previous  terms:
\begin{equation}
    \mL = \alpha \mathcal{L}_{\textrm{soft-IoU}} +   \mL_{\textrm{DA}}\;,
\end{equation}
where $\alpha$ is a hyper-parameter that controls their relative importance.

The model is trained by feeding it with both real and synthetic depth maps. However, it is worth pointing that 
the segmentation loss $\mathcal{L}_{\textrm{soft-IoU}}$ is only evaluated when the input is a synthetic depth map $\bX_i^s$, for which we do have the ground truth labels $\bM_i^s$. The real depth maps $\bX_i^r$  are considered only in the DA loss $\mL_{\textrm{DA}}$, which does not require ground truth segmentation labels.

\begin{figure*}[t!]
    \centering
    \includegraphics[width=\linewidth]{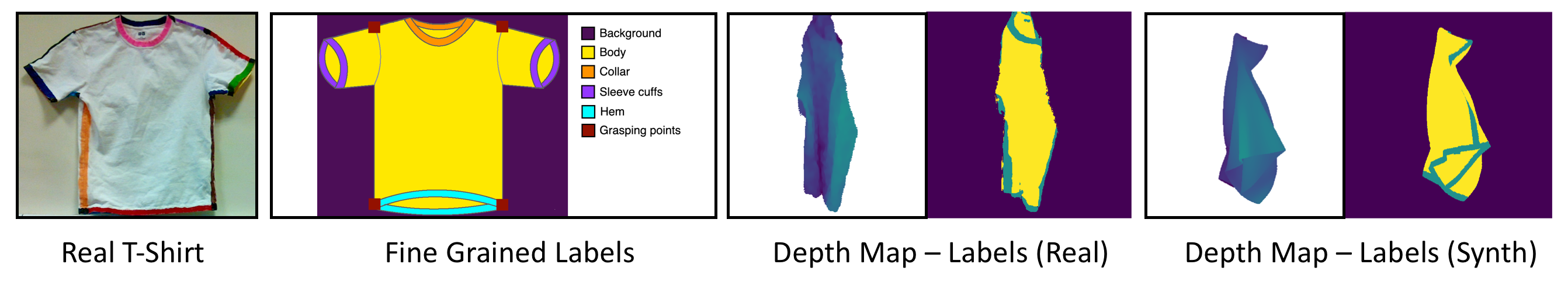}
    \caption{From left to right: T-shirt used for collection of real domain data; Fine-grained labels considered in our experiments; Depth map and segmentation labels for the real and synthetic domains ( edges are merged and labelled in {\em green}, main body in {\em yellow} and background in {\em dark blue}).}
    \label{fig:Painted-tshirt}
\end{figure*}



\section{DATASETS} \label{sec:dataset}
In this section we describe the process of collecting the real and generating the synthetic datasets, as well as the annotation procedure for each of them. 

\subsection{Synthetic Domain}
We show in Fig.~\ref{fig:Painted-tshirt} an example generated by using the physics cloth engine of Blender~\cite{Blender}.  The setup for generating the depth maps
consists in a deformed T-shirt model surrounded by a rig of $36$ cameras separated by steps of $10$ degrees around the cloth.  Specifically, the bounding box defined by the deformed mesh lies at the center of the circle, and we set the radius to $120cm$ to ensure the whole T-shirt mesh is completely visible by all cameras.
A 3D human body design suite~\cite{Makehuman} is used to obtain the T-shirt model. This model is defined by a quad mesh with $3.500$ vertices, which is the best topology for the cloth physics engine simulator. 
The cloth physics engine is based on a spring mass model, with  several cloth fabric presets and several parameters that are tunable for adjusting the behaviour of the simulation. 
We use the \textit{cotton} preset in the case of the T-shirt, and just modify the bending and stiffness parameters to achieve more realistic deformations.
The T-shirt mesh is hung from  a point and let deform by gravity. The deformation process is run for $250$ steps on each physics simulation to ensure a rest position is achieved. Before running each simulation, the mesh is randomly rotated, and a vertex is also randomly chosen as a hanging point. 
We split the data into 5600 training and 1500 test samples. Note that the test samples come from different hanging configurations (hanging point) than the training samples. The images in consecutive frames are  similar which would lead to a bias if random splits were used for training and testing. For the grasping point regression task, that shall be described in the experimental section, we use 2737 training and 344 test samples as not all examples have visible grasping points.

We also carry on a normalisation of the synthetic depth maps in the vertical and horizontal dimensions. For this purpose, the covariance matrices of all of the non-background pixels coordinates were averaged. The same matrix was obtained for the real dataset. The eigenvalues $\lambda$ from the singular value decomposition were utilised to scale the synthetic images. $\frac{\sqrt{\lambda^{r}}}{\sqrt{\lambda^{s}}}$   were used to scale the synthetic images along the vertical and horizontal axis to ensure a similar shape between real and synthetic depth maps. 

\subsection{Real Domain}




To facilitate the annotation of the real data, the T-shirt was first painted along the edges with different colours representing different edges and 4 grasping points as illustrated in Fig.~\ref{fig:Painted-tshirt}-left. 
The T-shirt is grasped by a Baxter robot and naturally hung under gravity. 
We manually adjusted real-world setup to roughly match the appearance and dynamics of the simulation. Specifically, an Intel RealSense L515 camera is placed $120cm$ away from the grasping point. The robot rotates the T-shirt every 10 degrees and depth images are captured.
The pseudo labels annotation was done by training a U-Net segmentation model from a small number of manually annotated examples with aggressive augmentation to adjust for the class imbalance. This was then applied to the rest of the real data and achieved the accuracy of 0.931 of mean intersection over union metric.

The depth images were min-max normalised after removing the background with a threshold in depth. Data augmentation techniques such as random horizontal flipping, rotation and cropping were utilised to enlarge the dataset. 
One dataset includes 1475 examples with 4 classes, i.e., all edges are considered as one class, then background and grasping points. 
Another datasets has 504 examples with 6 fine-grained labels i.e., bottom, top and middle edges, background and grasping points. We use 72 real test samples for ablation experiments which are two grasping point configurations $2\times36$.
For the grasping point regression task we use 300 training and 50 test samples as not all examples have visible grasping points.


\section{EXPERIMENTS}
 In this section we present the experimental results. We first report the results for segmentation of edges and grasping points. We then present the results for fine grained segmentation and a comparison to the state of the art.
\subsubsection{Main Results}


The results for edge and grasping point segmentation are shown in Table~\ref{table:result 4 class}. We report the performance for several labels that include background, body, edges and grasping points (GP) reported both in pixels and cm. The GP accuracy is measured as the average distance of the most confident prediction to the closest ground truth grasping point. The accuracy and precision of the background, body, and edges is measured via the IoU. 
For comparison as a simple baseline predictor of the grasping point we consider the centre of the cloth as well as a random grasping point. After the normalisation of the synthetic data the average distance from the cloth centre to its closest grasping point is $17cm$ (59.8pix) for the real dataset, and $11.7cm$  (41pix) for the synthetic. The average distance between randomly sampled points on the cloth to its closest grasping point is $11cm$ (39pix) for real dataset, and $10.4cm$ (36.4pix) for synthetic dataset. We evaluate only for one grasping point as in practice once the cloth is grasped by the predicted point, the model can be applied again to detect the next point.


   \begin{figure*}[t!]
      \centering
      \includegraphics[scale=0.55]{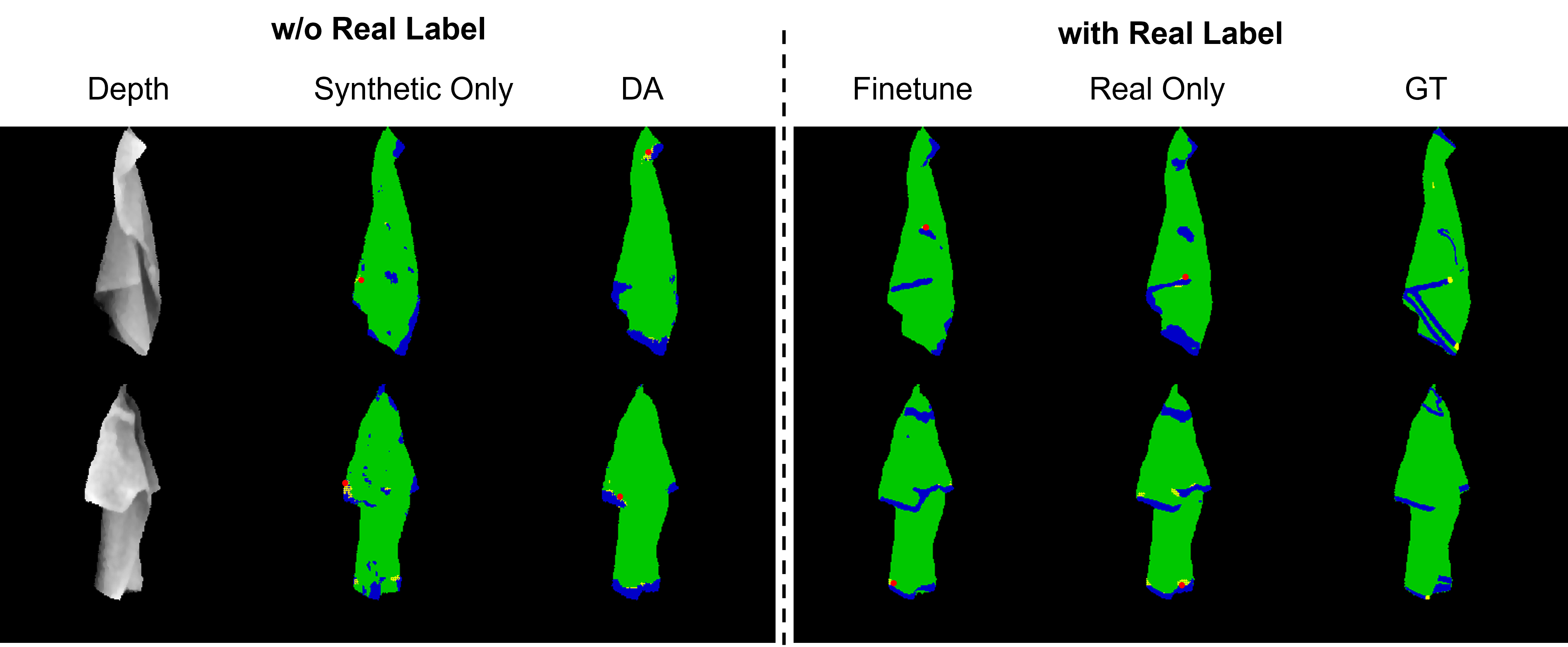}
      \caption{Visualization of the results, where background, cloth body, edges are denoted in \textit{Black}, \textit{Green}, \textit{Blue} respectively. Grasping points with highest confidence are highlighted by dots in \textit{Red}. }
      \label{result}
   \end{figure*}

As reported in Table~\ref{table:result 4 class} the performance of the body segmentation is nearly perfect, which is expected from the depth data, however the performance for edges is significantly lower. This is due to the complex folding of hanging cloth and overlapping of edges and body fabrics. It shows how challenging the edge segmentation task is. Note that except for Synth/Synth, the testing is done on real data only for all other evaluated methods. As expected within the domain train and test performance is best i.e.  Synth/Synth and finetuning also proves effective. More interestingly, the IoU results for our proposed DA are nearly twice better than for the direct synthetic to real model (Synth/Real). This is a clear indication that the proposed adversarial training is effective in bridging the gap between synthetic and real data.
Further observations can be made from the results for the grasping point detection. The best performance is obtained by our DA approach, followed by the same domain train/test experiments.  We believe this is due the grasping points benefiting more from training with large synthetic data as locally for GP the real and synthetic examples are more similar than globally on elongated edges.

\begin{table}[t!]
\resizebox{\textwidth}{!}{%
\begin{tabular}{ccccccl}
\Xhline{2\arrayrulewidth}
Train / Test & Background & Body & Edges & GP dist & GP dist\\
& (IoU) $\uparrow$ & (IoU) $\uparrow$ & (IoU) $\uparrow$ & (pix) $\downarrow$ & (cm) $\downarrow$ \\ 
\Xhline{2\arrayrulewidth}
 Synth / Synth & 0.999 & 0.888 & 0.481 & 23.76 & 6.78
 \\
Synth / Real & 0.997 & 0.857 & 0.133 & 36.23 & 10.35
 \\
Real / Real & 0.997 & 0.917 & 0.365  &   25.15 & 7.18
 \\ 
 Finetune & 0.998 & 0.923 &  0.381  & 38.0 & 10.85
 \\
 DA + Finetune & 0.984 & 0.867 & 0.362   & 32.76 & 9.36
  \\ 
 DA & 0.996 & 0.885 & 0.255 & 22.27 & { 6.36}
 \\ \Xhline{2\arrayrulewidth}
\end{tabular}%
}
\caption{Edge and grasping point results, where GP dist denotes the distance between the most confident predicted grasping point and the closest ground truth grasping point. Note that the distance is measured in pixels and the image size is 256$\times$256 pixels. Semantic segmentation performance is measured by Intersection over Union (IoU).}
\label{table:result 4 class}
\end{table}

In Table~\ref{table:result 6 class} we report the results for fine-grained segmentation, that includes additional top, bottom, middle labels for edges as well as grasping points. 
The performance for the edges drops compared to Table~\ref{table:result 4 class} as the model struggles to discriminate between bottom, middle and top edges in particular. Due to confusion between the labels, the IoU drops further as IoU penalises false positives and false negatives. For the grasping points the Real/Real is not significantly better than the Synth/Real, this is due to the limitation in the size of the real dataset. The Finetune greatly improves the quantitative performance as the model is able to extract more features via the pretrained weights. More importantly, the accuracy of detecting the grasping points by the proposed DA is nearly twice better than for training with synthetic or real only.

\begin{table}[t!]
\resizebox{\textwidth}{!}{%
\begin{tabular}{ccccccl}
\Xhline{2\arrayrulewidth}
Train / Test & Backgr. & Body & Top & Middle & Bottom & GP dist\\ 
& (IoU) $\uparrow$ & (IoU) $\uparrow$ & (IoU) $\uparrow$ & (IoU) $\uparrow$ & (IoU) $\uparrow$ & (cm) $\downarrow$ \\ 

\Xhline{2\arrayrulewidth}
 Synth / Synth & 0.999 & 0.929 & 0.278 & 0.560 & 0.567 & 3.89
\\
Synth / Real & 0.997 & 0.925 & 0.04 & 0.107 & 0.153 & 13.6
 \\
Real / Real&  0.997 & 0.918 & 0.09 & 0.251 & 0.215 & 13.1
 \\ 
 Finetune & 0.998 & 0.920 & 0.04 & 0.339 & 0.238 & 9.3 
 \\
 DA + Finetune & 0.998 & 0.927 & 0.03  & 0.290 & 0.270 &  7.9
 \\ 
 DA & 0.997 & 0.880 & 0.05 & 0.223 & 0.220 & 7.4
  \\ \Xhline{2\arrayrulewidth}
\end{tabular}%
}
\caption{Fine-grained results for 6 class labels. The performance for the edges drops significantly as the model struggles to discriminate between top, bottom, and side edges. However, the accuracy of detecting the grasping points by DA is nearly twice better than for training with synthetic or real only.}
\label{table:result 6 class}
\end{table}

\subsubsection{Other Results}
Table~\ref{table:result regression} shows the comparison to the closely related state of the art grasping point regression methods. Our DA classification model is trained on 5600 synthetic examples and gives the accuracy to 6.36cm.
 For comparative experiments we have also implemented a regression network based on the architecture proposed  in~\cite{Fan2020}, which we denote with \cite{Fan2020}$^+$. It obtained the best performance when training on 300 real samples and it is comparable to the results reported in \cite{Saxena-SII-2019} and \cite{Fan2020}$^\circ$, where 60.000 synthetic samples were used without real data.
 Note that the results for  \cite{Fan2020}$^\circ$,  \cite{Fan2020}$^\ast$ and \cite{Saxena-SII-2019} are for different and much larger datasets in the original papers, while  \cite{Fan2020}$^+$ and DA are for our dataset. Much better performance was achieved in~\cite{Fan2020}$^\ast$ by combining 60.000 synthetic and 500 real training samples. However, their dataset was less challenging, as the cloth was hung on a hanger, and the grasping point was always close to the collar.

 

\begin{table}[t!]
\resizebox{\textwidth}{!}{%
\begin{tabular}{ccccc}
\Xhline{2\arrayrulewidth}
 Zhang\cite{Fan2020}$^\ast$  &   Zhang\cite{Fan2020}$^\circ$ &   Zhang\cite{Fan2020}$^+$  & Saxena\cite{Saxena-SII-2019} & DA \\ \Xhline{2\arrayrulewidth}
 3.57 & 9.95 & 10.25 & 10 & { 6.36} \\
 \Xhline{2\arrayrulewidth}
 \end{tabular}%
}
\caption{Regression of grasping point (accuracy in cm). For comparison the median distance of the centre of the cloth to the nearest GP  is $17cm$ and the medians distance between a random point and the closest grasping point is $11cm$, for the real domain. }
\label{table:result regression}
\end{table}

 Finally we show some qualitative results in Fig.~\ref{result} for our model trained without (left) and with (right) the labels in real domain. Using the real labels for fine tuning visibly improves the accuracy of the segmentation. The real data allows the model to be more precise with the richer features extracted by the model pretrained with the synthetic dataset. The DA method also reliably predicts the edges, as it aids the model to better match the style of the features in both the two domains and thus improves on the model's ability generalise. The Finetune enhances the network's ability to be more precise in its predictions as it detects more edges than without finetuning.

\section{CONCLUSIONS}
In this paper we have investigated the problem of segmenting depth maps of highly deformed clothes into semantic regions that are useful for subsequent downstream tasks such as cloth grasping manipulation for folding, assisted dressing etc. We made two main contributions. First, the proposed architecture allows to predict regions of different type and extent, from local grasping points to larger semantic edges, without the need of retraining. Second, we devise a learning methodology that makes it possible to train the network using only semantic annotations on synthetic data, and addresses the domain gap between real and synthetic depth maps via a multi-layered domain adaptation strategy. The experiments show very promising results, in which the network trained with only synthetic ground truth annotations yields results on a par with a network trained with real  data.








\bibliographystyle{IEEEtran}
\bibliography{ICRA2022}

\end{document}